# Optimize_Prime@DravidianLangTech-ACL2022: Emotion Analysis in Tamil

**Omkar Gokhale**[*], **Shantanu Patankar**[*], **Onkar Litake**[†], **Aditya Mandke**[†], **Dipali Kadam**

Pune Institute of Computer Technology, Pune, India

omkargokhale2001@gmail.com, shantanupatankar2001@gmail.com,
onkarlitake@ieee.org, adeetya.m@gmail.com, ddkadam@pict.edu

## Abstract

This paper aims to perform an emotion analysis of social media comments in Tamil. Emotion analysis is the process of identifying the emotional context of the text. In this paper, we present the findings obtained by Team Optimize_Prime in the ACL 2022 shared task "Emotion Analysis in Tamil." The task aimed to classify social media comments into categories of emotion like Joy, Anger, Trust, Disgust, etc. The task was further divided into two subtasks, one with 11 broad categories of emotions and the other with 31 specific categories of emotion. We implemented three different approaches to tackle this problem: transformer-based models, Recurrent Neural Networks (RNNs), and Ensemble models. XLM-RoBERTa performed the best on the first task with a macro-averaged f1 score of 0.27, while MuRIL provided the best results on the second task with a macro-averaged f1 score of 0.13.

## 1 Introduction

Due to the rise in social media, internet users can voice their opinion on various subjects. Social networking platforms have grown in popularity and are used for a variety of activities such as product promotion, news sharing, and accomplishment sharing, among others (Chakravarthi et al., 2021). Emotion analysis or opinion mining is the study of extracting people's sentiment about a particular topic, person, or organization from textual data. Emotion analysis has many modern-day use-cases in e-commerce, social media monitoring, market research, etc. Tamil is the 18th most spoken language globally (Wikipedia contributors, 2022), with over 75 million speakers. Developing an approach for emotion analysis of Tamil text will benefit many people and businesses.

Emotion Analysis, at its core, is a text classification problem. To date, various approaches have been developed for text classification. Earlier, classification models like logistic regression, linear SVC, etc., were used. RNN based approaches like LSTMs also gained much traction because they produced better results than standard machine learning models. The introduction of transformers (Vaswani et al., 2017) changed the course of text classification due to their consistent performance. Multiple variations of the transformer have been developed like BERT (Devlin et al., 2018), AlBERT (Lan et al., 2019), XLM-RoBERTa (Conneau et al., 2019), MuRIL (Khanuja et al., 2021), etc.

In this paper, we have tried various approaches to detect emotions from social media comments. We have used three distinct ways to get optimal results: Ensemble models, Recurrent Neural Networks (RNNs), and transformer-based approaches. This paper will contribute towards future research in emotion analysis in low-resource Indic languages.

## 2 Related Work

Emotion Analysis has recently gained popularity, as large volumes of data are added to social networking sites daily. Earlier studies focus more on lexicon-based approaches, and they make use of a pre-prepared sentiment lexicon to classify the text. e.g., in Tkalčič et al. (2016), Wang and Pal (2015) and yan Nie et al. (2015), lexicon-based approaches are used; however, if unrelated words express emotions, this approach fails.

To overcome the limitations of lexical/keyword-based approaches, learning-based approaches were introduced. In this, the model learns from the data and tries to find a relationship between input text and the corresponding emotion. Researchers have tried out both supervised and unsupervised learning approaches. e.g., in Wikarsa and Thahir (2015), tweet classification was performed using naïve Bayes (supervised learning). In Hussien et al. (2016), SVM and multimodal naïve Bayes were used to classify Arabic tweets.

---

[*]first author, equal contribution

[†]second author, equal contribution

| Emotion | Neutral | Joy | Ambiguous | Trust | Disgust | Anticipation | Anger | Sadness | Love | Surprise | Fear |
|---|---|---|---|---|---|---|---|---|---|---|---|
| % | 34.24% | 15.3% | 11.82% | 8.6% | 6.3% | 5.9% | 5.7% | 5.07% | 4.8% | 1.63% | 0.7% |

Table 1: Class-wise distribution of data.

A combination of lexicon-based and learning-based approaches were used to perform classification on a multilingual dataset in Jain et al.(2017). Transfer learning-based approaches work well for low-resource languages. Transfer learning allows us to reuse the existing pre-trained models. For example, Ahmad et al.(2020) used a transfer learning approach to classify text in Hindi.

Lately, transformer-based models have been consistently outperforming other architectures, including RNNs. The development of models like MuRiL (Khanuja et al.,2021), XLM-RoBERTa (Conneau et al.,2019), Indic BERT (Kakwani et al.,2020), and M-BERT (Devlin et al.,2018) has encouraged research in various low resource as well as high resource languages.

## 3 Dataset Description

The shared task on Emotion Analysis in Tamil-ACL 2022 aims to classify social media comments into categories of emotions. The Emotion Analysis in Tamil Dataset (Sampath et al.,2022) consists of two datasets. The first dataset is for task A and has 11 categories of emotions which are: Neutral, Joy, Ambiguous, Trust, Disgust, Anger, Anticipation, Sadness, Love, Surprise, Fear. While the second is for task B and has 31 more specific categories of emotions. The distribution of data among classes is given in Table 1

### 3.1 Task A

The train, dev, and test datasets have 14,208, 3,552, and 4,440 data points, respectively. Each data point in the training data has the text in Tamil and its corresponding label in English.

### 3.2 Task B

The train, dev, and test datasets have 30,180, 4,269, and 4,269 data points, respectively. Each data point in the training data has the text in Tamil and its corresponding label, also in Tamil.
There is a significant class imbalance in the dataset, representing social media comments in real life.

## 4 Methodology

To classify social media comments into different emotions, we used three different approaches: ensemble models, Recurrent Neural Networks, and transformers. Figure 1. shows the architecture of all the three approaches[1].

### 4.1 Data Processing

#### 4.1.1 Data cleaning

We removed punctuations, URL patterns, and stop words. For better contextual understanding, we replaced emojis with their textual equivalents. For example, the laughing emoji was replaced by the Tamil equivalent of the word laughter.

Data cleaning boosted the performance of all RNN models and all transformer models except for MuRIL. MuRIL and all ensemble models worked best without data cleaning.

#### 4.1.2 Handling data imbalance

There is a significant class imbalance in the data. To reduce the imbalance, we used the following techniques: over-sampling, over-under sampling, Synthetic Minority Over-sampling (SMOTE) (Chawla et al.,2002), and assigning class weights. In over-under sampling, we under-sample the classes having more instances than expected and over-sample those having lesser instances than expected while keeping the length of the dataset constant. Over-under sampling worked best for all transformer and ensemble models, but it reduced the performance of RNN models. Assigning class weights to the input boosted the performance of the M-BERT - Logistic Regression ensemble model.

### 4.2 Ensemble model

As shown in the figure, we concatenate different machine learning models with multilingual BERT(M-BERT) (Devlin et al.,2018). Multilingual BERT is a BERT-based transformer trained in 104 languages. It simultaneously encodes knowledge of all these languages. M-BERT generates a sentence embeddings vector of length 768, with context. We then pass these embeddings to different machine learning models like logistic regression, decision trees, and XGBoost. We used grid search with macro-averaged f1 score as the scoring parameter for 3-5 cross-validation folds to fine-tune the hyperparameters.

---
[1]https://github.com/PICT-NLP/Optimize_Prime-DravidianLangTech2022-Emotion_Analysis

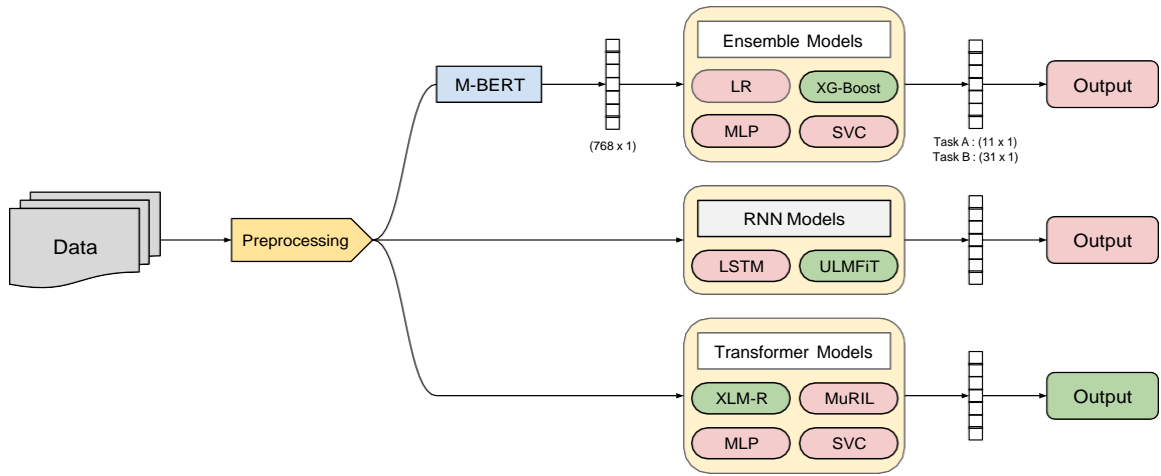

Figure 1: Model architecture (green box represents the classifier with highest f1 score in the group)

### 4.3 RNN Models

We have used two RNN models, Long Short-Term Memory(LSTM) networks and ULM-Fit.

#### 4.3.1 Vanilla LSTM

For setting a baseline for an RNN approach, we built word embeddings from scratch by choosing the top 64,000 most frequently occurring words in the dataset. This is passed through an embedding layer to get 100 dimension word vectors. The rest of the model includes a spatial drop out of 0.2, followed by the classification model consisting of two linear layers followed by a softmax.

#### 4.3.2 ULM-Fit

In transfer learning approaches, models are trained on large corpora, and their word embeddings are fine-tuned for specific tasks. In many state-of-the-art models, this approach is successful (Mikolov et al., 2013). Although Howard and Ruder (2018) argue that we should use a better approach instead of randomly initializing the remaining parameter. They have proposed ULMFiT: Universal Language Model Fine-tuning for Text Classification.

We use team gauravarora's (Arora, 2020) open-sourced models from the shared task at HASOC-Dravidian-CodeMix FIRE-2020. They build corpora for language modeling from a large set of Wikipedia articles. These models are based on the Fastai (Howard and Gugger, 2020) implementation of ULMFiT. We fine-tuned the models on Tamil, codemix datasets individually and on the Tamil-codemix combined dataset.

For tokenization, we used the Senterpiece module. The language model is based on AWD-LSTM (Merity et al., 2018). The model consists of a regular LSTM cell with spatial dropout, followed by the classification model consisting of two linear layers followed by a softmax.

### 4.4 Transformer Models

Our data sets consist of Tamil and Tamil-English codemixed data; we use four transformers MuRIL, XLM-RoBERTa, M-BERT, and Indic BERT. MuRIL (Khanuja et al., 2021) is a language model built explicitly for Indian languages and trained on large amounts of Indic text corpora. XLM-RoBERTa (Conneau et al., 2019) is a multilingual version of RoBERTa (Liu et al., 2019). Moreover, it is pre-trained on 2.5 TB of filtered CommonCrawl data containing 100 languages. M-BERT (Devlin et al., 2018) or multilingual BERT is pre-trained on 104 languages using masked language modeling (MLM) objective. Indic BERT (Kakwani et al., 2020) is a multilingual ALBERT (Lan et al., 2019) model developed by AI4Bharat, and it is trained on large-scale corpora of major 12 Indian languages, including Tamil. We use HuggingFace (Wolf et al., 2019) for training with SimpleTransformers. The training was stopped early if the f1 score did not improve for three consecutive epochs. A warning was given while training XLM-RoBERTa on the task B dataset using SimpleTransformers, which caused a considerable dip in the score obtained. The solution to this is to make the argument use_multiprocessing equal to False.

| Task A | | | |
|---|---|---|---|
| | Classifier | mf1 | wf1 |
| Ensemble Models | LR | 0.23 | 0.32 |
| | SVC | 0.18 | 0.33 |
| | XGBoost | 0.16 | 0.33 |
| | MLP | 0.19 | 0.32 |
| RNN Models | ULMFIT | 0.27 | **0.41** |
| | LSTM | 0.21 | 0.33 |
| Transformer Models | MuRIL | 0.31 | 0.37 |
| | XLM-R | **0.32** | 0.37 |
| | M-BERT | 0.27 | 0.36 |
| | IndicBERT | 0.29 | 035 |

Table 2: Results of task A
(mf1: macro avg f1, wf1: weighted avg f1)

| Task B | | | |
|---|---|---|---|
| | Classifier | mf1 | wf1 |
| Ensemble Models | LR | 0.10 | 0.17 |
| | SVC | 0.09 | 0.20 |
| | XGBoost | 0.07 | 0.17 |
| | MLP | 0.08 | 0.17 |
| RNN Models | LSTM | 0.11 | **0.21** |
| Transformer Models | MuRIL | **0.13** | 0.16 |
| | IndicBERT | 0.09 | 0.11 |

Table 3: Results of Task B
(mf1: macro avg f1, wf1: weighted avg f1)

## 5 Results

The results obtained for Task A and Task B are given in Table 2 and Table 3, respectively.

### 5.1 Ensemble models

In task A, logistic regression achieved the best results with macro-averaged f1 scores of 0.23. MLP achieved a macro averaged f1 score of 0.19. Support Vector Machine also produced decent results with a macro-averaged f1 score of 0.18 and a weighted-average f1 score of 0.33.
For task B, logistic regression got a macro average f1 score of 0.1 and outperformed all the other ensemble models.

### 5.2 RNNs

For task A, ULMFit performed well with a macro-averaged f1 score of 0.27. For task B, LSTM generated a macro-averaged f1 score of 0.11 and a weighted-average f1 score of 0.21.

### 5.3 Transformers

For task A, XLM-RoBERTa outperformed all other models with a macro averaged f1 score of 0.32 and a weighted-average score of 0.37. Performance of MuRIL was similar to XLM-Roberta. For task B, MuRIL outperformed all other models with a macro-averaged f1 score of 0.125.

Overall, XLM-RoBERTa performed the best on Task A(11 classes) while MuRIL performed the best on Task B(31 labels)

## 6 Conclusion

The aim of this paper was to classify social media comments. We used three approaches: Ensemble models, Recurrent Neural Networks (RNNs), and transformers. Out of these models, for task A, XLM-RoBERTa outperformed all other models with a macro-averaged f1 score of 0.27. However, in Task B, MuRIL outperformed all other models with a macro averaged f1 score of 0.125. Overall, it is observed that the models classify emotions like Joy, Sadness, Neutral, and sentences having ambiguity well. However, the models classify more complex emotions like anger, fear, and sadness with much less accuracy. In the future, various techniques like genetic algorithm-based ensembling can be tried to improve the performance of the models.

## 7 Acknowledgments

We want to thank SCTR's Pune Center for Analytics with Intelligent Learning for Multimedia Data for their continuous support. A special thanks to Neeraja Kirtane and Sahil Khose for their help in drafting the paper.